# Automated Intracranial Artery Labeling using a Graph Neural Network and Hierarchical Refinement

Li Chen[1][0000-0003-0233-4576], Thomas Hatsukami[2], Jenq-Neng Hwang[1], Chun Yuan[3*]

`{cluw,tomhat,hwang,cyuan}@uw.edu`

[1] Department of Electrical and Computer Engineering, University of Washington, Seattle, WA, USA

[2] Department of Surgery, University of Washington, Seattle, WA, USA

[3] Department of Radiology, University of Washington, Seattle, WA, USA

**Abstract.** Automatically labeling intracranial arteries (ICA) with their anatomical names is beneficial for feature extraction and detailed analysis of intracranial vascular structures. There are significant variations in the ICA due to natural and pathological causes, making it challenging for automated labeling. However, the existing public dataset for evaluation of anatomical labeling is limited. We construct a comprehensive dataset with 729 Magnetic Resonance Angiography scans and propose a Graph Neural Network (GNN) method to label arteries by classifying types of nodes and edges in an attributed relational graph. In addition, a hierarchical refinement framework is developed for further improving the GNN outputs to incorporate structural and relational knowledge about the ICA. Our method achieved a node labeling accuracy of 97.5%, and 63.8% of scans were correctly labeled for all Circle of Willis nodes, on a testing set of 105 scans with both healthy and diseased subjects. This is a significant improvement over available state-of-the-art methods. Automatic artery labeling is promising to minimize manual effort in characterizing the complicated ICA networks and provides valuable information for the identification of geometric risk factors of vascular disease. Our code and dataset are available at https://github.com/clatfd/GNN-ART-LABEL.

**Keywords:** Artery Labeling, Graph Neural Network, Hierarchical Refinement, Intracranial Artery.

## 1    Introduction

Intracranial arteries (ICA) have complex structures and are critical for maintaining adequate blood supply to the brain. There are substantial variations in these arteries among individuals that are associated with vascular disease and cognitive functions [1–3]. Comprehensive characterization of ICA including labeling each artery segment with its anatomical name (Fig. 1 (b)) is desirable for both clinical evaluation and research. The center of the ICA is the Circle of Willis (CoW, normally incorporating nine artery segments forming a ring shape), which connects the left and right hemispheres, as well as anterior and posterior circulations. It has been reported that only 52% of the population has a complete CoW [4]. Many natural variations of CoW exist, including those missing one or multiple arterial segments [5]. In addition, disease related changes within the complex network of ICA are also challenging for automated labeling. For example, stenosis may cause decreased cerebral flow, reflected as reduced blood signal in arterial images; collateral flow forms near the severe stenosis, leading to abnormal structures in the ICA. These situations make automated artery labeling challenging. A simplified graph illustration of ICA is shown in Fig. 1 (c).

---

* Corresponding author

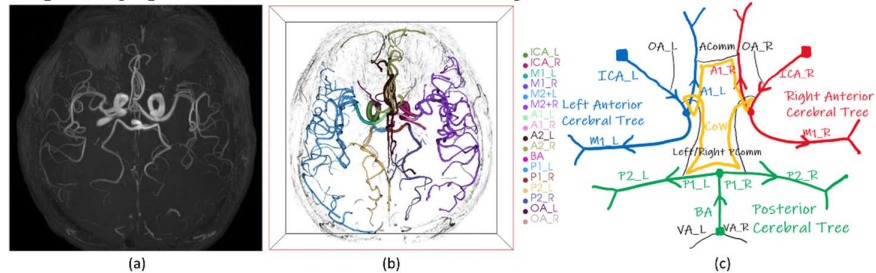

**Fig. 1.** (a) Time of flight (ToF) Magnetic Resonance Angiography (MRA) of cerebral arteries. (b) ICA labeled in different colors. (c) Illustration of CoW (yellow), left (blue) and right (red) anterior circulation, posterior circulation (red) and optional artery branches (black) with their anatomical names. When there are ICA variations, not only the optional artery branches, but also A1, M1, P1 segments may be missing. See supplementary material for abbreviations.

There have been continuous efforts in automating ICA labeling, using either private datasets with a limited number of scans or the publicly available UNC dataset with 50 cerebral Magnetic Resonance Angiography (MRA) images [6]. Takemura et al. [7] built a template of the CoW on five subjects, then arteries were labeled by template alignment and matching on fifteen scans. A more complete artery atlas was built from a population-based cohort of 167 subjects by Dunås et al. [8, 9] using a similar matching approach, and arteries were labeled in 10 clinical cases. Bilgel et al. [10] considered connection probability within the cerebral network using belief propagation for labeling 30 subjects but the method was limited to anterior circulations. Using the UNC dataset, in the serial work from Bogunović et al. [11–13], eight typical ICA graph templates were used to represent ICA with variations, and bifurcations of interest (BoI) were defined and classified so that vessels were labeled indirectly. However, more variations exist beyond the eight typical types. Using the same dataset, by combining artery segmentation along with the labeling, Robben et al. [14] simultaneously optimized the artery centerlines and their labels from an over complete graph. However, their computation involved thousands of variables and constraints, and takes as long as 510 seconds per case. In summary, while previous works have shown success in labeling relatively small datasets with limited variations in mostly healthy populations, prior knowledge about the global artery structures and relations has not been fully explored.

Furthermore, labeling efficiency has not been considered for a large number of scans, which will be needed for clinical applications.

The Graph neural network (GNN) is an emerging network structure recently attracting significant interest [15, 16], including applications on vasculature [17, 18]. By passing information between nodes and edges within the graph, useful properties for the graph can be predicted. Considering the graph topology in anatomical structures of ICA, in this work, we propose a GNN model with hierarchical refinement (HR), aiming to overcome the challenges in arterial labeling by training with large and diversely labeled datasets (more than 500 scans in the training set from multiple sources) and applying refinements after network predictions to combine prior knowledge on ICA. In addition to its superior performance compared with methods described in the literature, our work shows robustness and generalizability on various challenging anatomical variations.

## 2  Methods

### 2.1  Intracranial artery labeling

The definition of arteries and their abbreviations follows [19] (also in supplementary material). All visible ICA in MRA are traced and labeled using a validated semi-automated artery analysis tool [19, 20] by experienced reviewers with the same labeling criteria, then examined by a peer-reviewer to control quality. Arteries not connected in the main artery tree are excluded from labeling, such as the arteries outside the skull.

### 2.2  Graph neural network (GNN) for node and edge probabilities

The ICA network is represented as the centerlines of arteries, each with consecutively connected 3D points with radius. Centerlines in one MRA scan are constructed as an attributed relational graph $G = (V, E)$. $V = \{\boldsymbol{v}_i\}$ represents all unique points in the centerlines with node features of $\boldsymbol{v}_i$, and $E = \{\boldsymbol{e}_k, r_k, s_k\}$ represents all point connections where edge $k$ connects between the node index $r_k, s_k$ with edge features of $\boldsymbol{e}_k$. $e(i)_{1,\dots,D(i)}$ are all the edges connected with node $i$ ($r_k = i$ or $s_k = i$). $D(i)$ is the degree (number of neighbor nodes) of node $i$.

Features for node $\boldsymbol{v}_i$ include $\boldsymbol{p}_i$ for $x, y, z$ coordinates, $r_i$ for radius and $\boldsymbol{b}_i$ for the directional embedding of the node. Due to the uncertain number of edges connected to the node, direction features cannot be directly used as an input in GNN. Here we use the multi-label binary encoding to represent direction features. First, 26 major directions in the 3D space are defined as $n_u = (x_u, y_u, z_u)_{u=1,\dots,26}$, with 45 degrees apart in each axis, excluding duplicates.

$$\begin{cases} x = \sin(45° * a) * \cos(45° * b) \\ y = \cos(45° * a) * \cos(45° * b), \ \ a \in \{0, \dots, 7\}, b \in \{-2, \dots, 2\} \\ \quad\quad z = \sin(45° * b) \end{cases} \quad (1)$$

Then each edge direction $(x_v, y_v, z_v)$ originating from the node is matched with the major directions with $dir_v = argmax_u(x_u x_v + y_u y_v + z_u z_v)$. $\boldsymbol{b}_i$ is the 26-dimensional feature with encoded direction for all $dir_v$. (an example in supplementary figure)

Features for edges $\boldsymbol{e}_k$ include edge direction $\boldsymbol{n}_k = (\boldsymbol{p}_{s_k} - \boldsymbol{p}_{r_k})$, which is then normalized (and inverted) so that $||\boldsymbol{n}_k|| = 1$ and $z_k > 0$; distance between nodes at two ends $d_k = ||\boldsymbol{p}_{s_k} - \boldsymbol{p}_{r_k}||$; and mean radius at two nodes $\bar{r}_k = (r_{s_k} + r_{r_k})/2$.

With similar purpose of BoI [11–13], we remove all nodes with a degree of 2 to reduce the graph size, as nodes requiring labeling are usually at bifurcations or ending points. If the remaining nodes are correctly predicted as one of the 21 possible bifurcation/ending types, then the ICA (edges) can be labeled based on their connections.

We implemented the GNN based on the message passing GNN framework proposed in [16, 21] to predict the types for each node and edge. The GNN takes a graph with node and edge features as input and returns a graph as output with additional features for node and edge types. The input features of edges and nodes in the graph are encoded to an embedding in the encoder layer. Then the core layer passes messages for 10 rounds by concatenating the encoder's output with the previous output of the core layer. The embedding is restored to edge and node features in the decoder layer with additional label features. Computation in each graph block is shown in the equation 2. The edge attributes are updated through the per-edge "update" function $\emptyset^e$, and features for edges connected to the same node are "aggregated" through function $\rho^{e \rightarrow v}$ to update node features through the per-node "update" function $\emptyset^v$. The network structure is shown in Fig. 2.

$$\begin{cases} updated\ edge\ attributes\ \boldsymbol{e}'_k = \emptyset^e(\boldsymbol{e}_k, \boldsymbol{v}_{r_k}, \boldsymbol{v}_{s_k}) \\ updated\ edge\ attributes\ per\ node\ \bar{\boldsymbol{e}}_i{}' = \rho^{e \rightarrow v}(\{(\boldsymbol{e}'_k, r_k, s_k)\}_{r_k = i}) \\ updated\ node\ attributes\ \boldsymbol{v}'_i = \emptyset^v(\bar{\boldsymbol{e}}_i{}', \boldsymbol{v}_i) \end{cases} \quad (2)$$

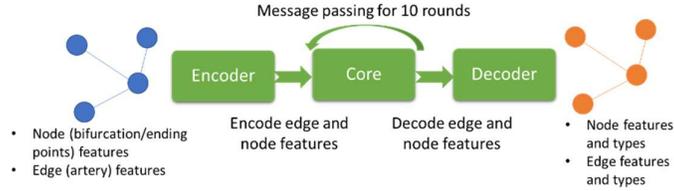

**Fig. 2.** GNN structure used in this study.

Probability $P_{nt}(i)$ for node $i$ being bifurcation/ending type $nt \in \{0: Non\_Type, 1: ICA\_Root\_L, \dots, 20: ICA\_PComm\}$ is calculated using a softmax function of GNN output $O_{nt}(i)$. The predicted node type $T_n(i)$ is then identified by selecting the node type with the maximum probability.

$$P_{nt}(i) = \frac{e^{O_{nt}(i)}}{\sum_{nti=0}^{20} e^{O_{nti}(i)}} \quad (3)$$

$$T_n(i) = argmax_{nt}(P_{nt}(i)) \quad (4)$$

Similar for edges, $et \in \{0: Non\_Type, 1: ICA\_L, ..., 22: OA\_R\}$, the edge probability and predicted edge type are

$$P_{et}(k) = \frac{e^{O_{et}(k)}}{\sum_{eti=0}^{22} e^{O_{eti}(k)}} \tag{5}$$

$$T_e(k) = argmax_{et}(P_{et}(k)) \tag{6}$$

Ground truth types for nodes and edges are $G(i), G(k)$.

The GNN was trained using combined weighted cross entropy losses in both nodes and edges, with weights inverse proportional to frequencies of the node and edge types. Batch size of 32 graphs was used in training the GNN. Adam optimizer [22] was used for controlling the learning rate. Positions of nodes from different datasets were normalized based on the imaging resolution, and a random translation of positions (within 10%) was used as the data augmentation method.

### 2.3 Hierarchical refinement (HR)

Predictions from the GNN might not be perfect, as end-to-end training cannot easily learn global ICA structures and relations. Human reviewers are likely to subdivide ICA into three sub-trees (i.e., left/right anterior, posterior cerebral trees), find key nodes (such as the bifurcation for ICA/MCA/ACA) in sub-trees, then add additional branches which are less important and more prone to variations (such as PComm, AComm). Enlighted by the sequential behavior during manual labeling, a hierarchical refinement (HR) framework based on GNN outputs is proposed to further improve the labeling. Starting from the most confident nodes, the three-level refinement is shown in Fig. 3.

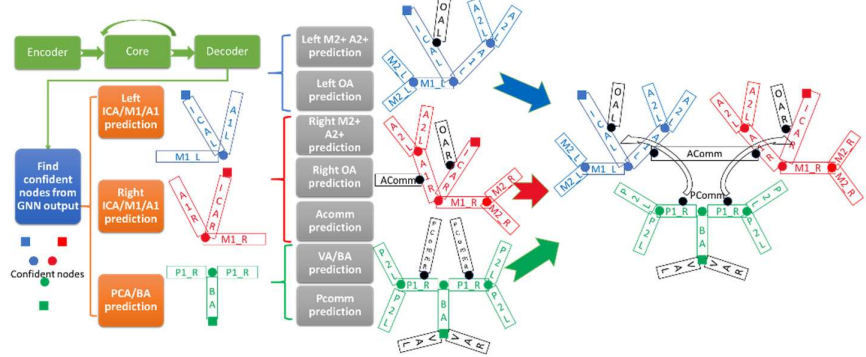

**Fig. 3.** Workflow of HR framework. In the first level (blue box), confident nodes (circle and square dots) are identified from the GNN outputs. In the second level (orange boxes), confident nodes as well as their inter-connected edges in the left (blue lines)/right (red lines) anterior, posterior (green) sub-trees are identified. In the third level (grey boxes), optional nodes and edges (black lines) are added to each of the three sub-trees to form a complete artery tree.

**Level one labeling.** We consider nodes as confident if the predicted node type fits the predicted edge types in edges they are connected with.

$$F\left(\{T_e\big(e(i)_{1,\ldots,D(i)}\big)\}\right) = T_n(i) \tag{7}$$

$F$ is a lookup table for all valid pairs of edge types and node types. For example, $F(P1\_L, P1\_R, BA) = PCA/BA$, $F(ICA\_L) = ICA\_ROOT\_L$.

**Level two labeling.** From confident nodes, three sub-trees are built, and major-branch nodes are predicted in each sub-tree individually. Major node $i$ is defined as ICA/MCA/ACA (for anterior trees) and PCA/BA (for posterior trees), and branch nodes $j$ are defined as ICA_Root, M1/2, A1/2 (for anterior trees) and BA/VA, P1/2 (for posterior trees). If major nodes are not confident nodes in each sub-tree, they are predicted with type $argmax_i\left(P_{nt_i}(i)|D(i) \neq 1\right)$ with additional constraints if branch nodes $j$ are confident ($i \notin j$ and $i$ must be in the path between any pair of $j$). Then from the major node, all unconfident branch nodes are predicted using the target function of

$$\begin{cases} argmax_j\left[P_{nt_j}(j) + P_{et_{e(i)}}\big(e(i)_{1,\ldots,D(i)}\big)_{r_{e(i)}=j, s_{e(i)}=i}\right], if\ P_{nt_i}(i) > Thres \\ argmax_j\left(P_{nt_i}(j)\right), if\ P_{nt_i}(i) < Thres \end{cases} \tag{8}$$

On rare occasions, when the major nodes have a probability lower than a certain threshold $Thres$ (when there are anatomical variations where major nodes do not exist), branch nodes are predicted without edge probability.

If certain distance between the optimal $i$ and $j$ is beyond the mean plus 1.5 standard deviation of $G(k)|r_k = j, s_k = i$ from the training set, labeling on $j$ will be skipped and a node with a degree of 2 will be labeled so that its distance to node $i$ is closest to the mean distance of $G(k)$. This happens when there are missing Acomm or Pcomm.

**Level three labeling.** Optional branches are added to three sub-trees. M2+, A2+, and P2+ edges are assigned for all distal neighbors of M1/2, A1/2, P1/2 nodes. Based on node probabilities, OAs are identified on the path between ICA_Root and ICA/MCA/ACA nodes, Acomm is assigned if there is a connection between A1/2_L and A1/2_R, Pcomm is assigned if there is connection between P1/2 and ICA/MCA/ACA, VA_Root is predicted from neighbors of BA/VA.

## 2.4 Experiments

**Datasets**. Five datasets from our previous research [23][23][23][23]were used to train and evaluate our method, then the generalizability was assessed on the public UNC dataset with/without further training. Details for the datasets are in supplementary material.

Our five datasets were collected with different resolutions from different scanner manufacturers. Subjects enrolled in the datasets include both healthy (no recent or chronic vascular disease) and with various vascular related diseases, such as recent stroke events and hypertension. All the datasets were randomly divided into a training set (508 scans), a validation set (116 scans) and a testing set (105 scans). If the subject had multiple scans, these scans were divided into the same set. All scans from the UNC dataset (https://public.kitware.com/Wiki/TubeTK/Data, healthy volunteers) with publicly available artery traces (N=41) were used. Generally, our dataset has more ICA variations and more challenging anatomies than the UNC dataset.

**Evaluation metrics.** As our purpose is to label the ICA, the accuracy of predicted node labels is the primary metric for evaluation (Node_Acc). In addition, we also used number of wrongly predicted nodes per scan (Node_Wrong), edge accuracy (Edge_Acc) and the percent of scans with CoW nodes (ICA/MCA/ACA, PCA/BA, A1/2, P1/2/PComm, PComm/ICA), all nodes and all edges correctly predicted (CoW_Node_Solve, Node_Solve, Edge_Solve). For detailed analysis of detection performance on each bifurcation type, the detection accuracy, precision and recall for 7 major bifurcation types (ICA-OA, ICA-M1, ICA-PComm, ACA1-AComA, M1-M2, VBA-PCA1, PCA1-PComA) were calculated. The processing time was also recorded. Due to the lack of criteria for labeling nodes with degrees of 2, nodes such as A1/2 without AComm were excluded from the evaluation.

**Comparison methods.** With the same artery traces of our dataset, three artery labeling methods [7, 9, 19] were used to compare the performance. Due to the unavailability of two methods using the UNC dataset, we only cite evaluation results from their publications. Direction features and HR were sequentially added to our baseline model to evaluate the contribution of different features and the effectiveness of the HR framework.

As the ablation study, GNN without HR predicts node and edge types directly from the GNN outputs $T_n(i)$ and $T_e(k)$. We further tested the removal of direction features.

## 3 Results

In the testing set of our dataset, 1035 confident nodes (9.86/scan) were identified, and 5 of them (0.5%, none are major or branch nodes) were predicted wrongly, showing labeling of confident nodes is reliable, so that labeling in the following up levels in HR was meaningful. *Thres* was chosen as 1e-10 from the validation set. Examples of correctly labeling challenging cases are shown in Fig. 4. Our method was robust, even with artificial noise branches added in the M1 branch shown in Fig. 4 (d).

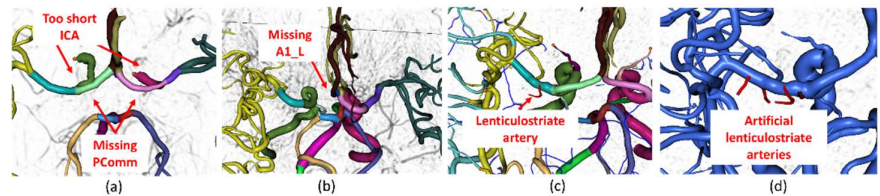

**Fig. 4** Examples of challenging anatomical variations where our method predicted all arteries correctly. (a) A subject with Parkinson's disease. Occlusions cause both right and left internal carotid arteries to be partially invisible. In addition, Pcomms are missing. (b) A hypertensive subject with rare A1_L artery missing, which is not among the 8 anatomical types and thus not solvable in [13]. (c) Some lenticulostriate arteries are visible in our dataset with higher resolution, an additional challenge for labeling, our method predicted it correctly as a non-type. (d) With more artificial lenticulostriate arteries added in the M1_L segment, our method is still robust to these additional noise branches.

**Table 1.** Comparison with existing methods and the ablation study on our testing set (N=105).

| Method | Node_Acc↑ | Node_Wrong↓ | Node_Solve↑ | CoW_Node_Solve↑ | Edge_Acc↑ | Edge_Solve↑ | Process time (s)↓ |
|---|---|---|---|---|---|---|---|
| MAP [19] | 0.9153 | 10.0 | 0 | 0.0476 | 0.3304 | 0 | 1.075 |
| Template [7] | 0.7316 | 31.6 | 0 | 0.0476 | 0.7934 | 0 | 5.057 |
| Atlas [9] | 0.8856 | 13.5 | 0 | 0.0095 | 0.7010 | 0 | 9.253 |
| GNN(Pos) | 0.9553 | 5.3 | 0.0286 | 0.3524 | 0.9099 | 0 | **0.017** |
| GNN(Pos+Dir) | 0.9637 | 4.3 | 0.0381 | 0.4286 | 0.9223 | 0 | 0.020 |
| GNN(Pos+Dir)+HR | **0.9746** | **3.0** | **0.3238** | **0.6381** | **0.9246** | **0.3238** | 0.092 |

The comparison with other artery labeling algorithms and the ablation study is shown in Table 1. Our method demonstrates a better node accuracy of 97.5% with 3.0 wrong nodes/scan. Our method is the only one with cases where all nodes and edges were predicted correctly with the minimum processing time (less than 0.1 seconds). With direction features and the HR added to the baseline model, the performance is further improved. ICA-OA is the most accurately detected bifurcation type with detection accuracy of 96.2% while the challenging M1/2 has an accuracy of 68.1%. Mean detection accuracy is 83.1%, precision is 91.3%, recall is 83.8%.

**Table 2.** Performance of detection accuracy (A), precision (P) and recall (R) for each bifurcation type, compared with previous methods using the UNC dataset. Note that our method is trained with our dataset with (GNN+HR+UNC) and without (GNN+HR) further training in the UNC dataset, but other methods were trained and evaluated by leave-one-out cross validation on the UNC dataset alone (more likely to overfit on the UNC dataset).

| Method | GNN+HR (Ours) | | | GNN+HR+UNC (Ours) | | | Robben [14] | | | Bogunović [13] | | |
|---|---|---|---|---|---|---|---|---|---|---|---|---|
| | A | P | R | A | P | R | P | R | R | A | P | R |
| ICA-OA | 97 | 100 | 97 | 100 | 100 | 100 | 99 | 99 | 100 | 99 | 100 | 99 |
| ICA-M1 | 90 | 96 | 93 | 95 | 97 | 97 | 100 | 100 | 100 | 99 | 99 | 100 |
| ICA-PComm | 88 | 97 | 89 | 95 | 97 | 97 | 98 | 100 | 98 | 93 | 94 | 96 |
| ACA1-AComA | 95 | 100 | 95 | 96 | 100 | 96 | 96 | 100 | 95 | 92 | 93 | 97 |
| M1-M2 | 90 | 95 | 95 | 95 | 97 | 97 | 78 | 78 | 100 | 89 | 89 | 100 |
| VBA-PCA1 | 94 | 97 | 97 | 90 | 94 | 94 | 90 | 98 | 91 | 94 | 100 | 93 |
| PCA1-PComm | 90 | 97 | 92 | 92 | 98 | 94 | 95 | 100 | 92 | 96 | 100 | 94 |
| Mean | 92 | **97** | 94 | **95** | **97** | **97** | 94 | 96 | **97** | **95** | 96 | **97** |

Our method showed good generalizability on the UNC dataset (Table 2). Even without additional training on the UNC dataset, the node accuracy was 99.03% (2.0 wrong nodes/scan) with 56% of cases solved. Mean detection accuracy for all node types was 92%. As a reference with methods using leave-one-out cross validation trained and evaluated using the same dataset, 58% of cases were solved [13]. The mean detection accuracies were 94% and 95% in [13, 14], respectively. If trained in combination with the UNC dataset using three fold cross validation, our method outperforms [13, 14].

## 4    Discussion and Conclusion

We have developed a GNN approach to label ICA with HR on our comprehensive ICA dataset (729 scans). Four contributions and novelties in our work are worth highlighting. 1) The dataset includes more diverse and challenging ICA variations compared with the existing UNC dataset, which is better suited to evaluate labeling performance. 2) The GNN and HR framework is an ideal method to learn from the graph representation of ICA and incorporate prior knowledge about ICA structure. 3) With accurate predictions of 20 node and 22 edge types covering all major artery branches visible in MRA, this method can automatically provide comprehensive features for detailed analysis of cerebral flow and structures in less than 0.1 seconds. 4) It should also be noted that our GNN and HR framework is not only applicable to ICA, but also to any graph structures where sequential labeling helps. For example, major lower extremity arteries and branches can be labeled ahead of labeling the collateral arteries.

Accurate ICA labeling using our method relies on reliable artery tracing, which is one of our limitations, although this is a lesser concern compared with non-graph based methods, where some artery tracing mistakes (such as centerlines off-center or zigzags in the path) can be avoided through a simplified representation of the ICA through graph constructions.

## 5    Acknowledgements

This work was supported by National Institute of Health under grant R01-NS092207. We are grateful for the collaborators who provided the datasets for this study, including the CROP and BRAVE investigators, and researchers from the University of Arizona, USA, Beijing Anzhen hospital, China, and Tsinghua University, China and the public data from The University of North Carolina at Chapel Hill (distributed by the MIDAS Data Server at Kitware Inc.). We acknowledge NIVIDIA for providing the GPU used for training the neural network model.

Our code and dataset are available at https://github.com/clatfd/GNN-ART-LABEL.